\documentclass[16pt]{article}

\usepackage{arxiv}

\usepackage[utf8]{inputenc} 
\usepackage[T1]{fontenc}    
\usepackage{hyperref}       
\usepackage{url}            
\usepackage{booktabs}       
\usepackage{amsfonts}       
\usepackage{nicefrac}       
\usepackage{microtype}      
\usepackage{lipsum}
\usepackage{graphicx}
\usepackage{amsmath}
\graphicspath{ {./images/} }

\title{DiffSOS: Acoustic Conditional Diffusion Model for Speed-of-Sound Reconstruction in Ultrasound Computed Tomography}

\author{
  Yujia Wu\textsuperscript{1} \quad
  Shuoqi Chen\textsuperscript{1} \quad
  Shiru Wang\textsuperscript{1} \quad
  Yucheng Tang\textsuperscript{2} \quad
  Petr Bruza\textsuperscript{1} \quad
  Geoffrey P. Luke\textsuperscript{1}\thanks{Corresponding author: geoffrey.p.luke@dartmouth.edu} \\
  \\
  \textsuperscript{1}Thayer School of Engineering, Dartmouth College, Hanover, NH, USA \\
  \textsuperscript{2}NVIDIA Corp \\
}

\begin{document}
\maketitle
\begin{abstract}
Accurate Speed-of-Sound (SoS) reconstruction from acoustic waveforms is a cornerstone of ultrasound computed tomography (USCT), enabling quantitative velocity mapping that reveals subtle anatomical details and pathological variations often invisible in conventional imaging. However, practical utility is hindered by the limitations of existing algorithms; traditional Full Waveform Inversion (FWI) is computationally intensive, while current deep learning approaches tend to produce oversmoothed results lacking fine details. We propose \textbf{DiffSOS}, a conditional diffusion model that directly maps acoustic waveforms to SoS maps. Our framework employs a specialized acoustic ControlNet to strictly ground the denoising process in physical wave measurements. To ensure structural consistency, we optimize a hybrid loss function that integrates noise prediction, spatial reconstruction, and noise frequency content. To accelerate inference, we employ stochastic Denoising Diffusion Implicit Model (DDIM) sampling, achieving near real-time reconstruction with only 10 steps. Crucially, we exploit the stochastic generative nature of our framework to estimate pixel-wise uncertainty, providing a measure of reliability that is often absent in deterministic approaches. Evaluated on the OpenPros USCT benchmark, DiffSOS significantly outperforms state-of-the-art networks, achieving an average Multi-scale Structural Similarity of 0.957. Our approach provides high-fidelity SoS maps with a principled measure of confidence, facilitating safer and faster clinical interpretation.
\end{abstract}


\section{Introduction}
Accurate reconstruction of acoustic properties from ultrasound data is critical for advanced medical imaging. Among these, the Speed-of-Sound (SoS) map serves as a vital quantitative biomarker, offering high-contrast information about tissue composition and pathology that is often invisible in standard B-mode ultrasound~\cite{rau2021speed}. For instance, it enables clinicians to quantify tissue density for the early detection of solid tumors. In Ultrasound Computed Tomography (USCT), the target anatomy is surrounded by a ring array of transducers immersed in an acoustic coupling medium. Data acquisition follows a synthetic aperture protocol; transducers emit ultrasonic pulses sequentially, while all or a subset of elements simultaneously record the resulting pressure fields~\cite{10793606}. This process yields a dense set of radiofrequency acoustic waveforms, time-series signals that encode complex wave-propagation physics and serve as the source data for reconstruction. The goal is to recover the underlying SoS map, a spatial distribution of local wave velocities that is directly related to tissue density and elasticity. However, translating raw acoustic waveforms into high-resolution SoS maps remains a challenging inverse problem. The gold standard is Full Waveform Inversion (FWI), which treats reconstruction as an iterative optimization of the wave equation~\cite{tarantola1984inversion}. However, FWI is computationally intensive and highly sensitive to initial velocity models~\cite{byrd1995limited}. Poor initialization often traps the optimization in local minima, leading to cycle-skipping artifacts that compromise clinical utility~\cite{virieux2009overview}.

Recent Deep Learning methods offer a faster alternative to iterative FWI. Deterministic regression models, such as U-Nets, allow for real-time SoS estimation~\cite{Wu_2020} but often suffer from regression to the mean, resulting in oversmoothed images that lack sharp structural boundaries~\cite{lee2024systematicbiasmachinelearning,7797130}. A critical limitation of existing learning-based frameworks is their reliance on precomputed proxies, such as Time-of-Flight maps or low-quality SoS maps~\cite{tof,liu2025reconstruction}. This imposes additional computational overhead and create an information bottleneck by discarding phase and diffraction data. While Generative Adversarial Networks (GANs) have been explored to recover texture~\cite{ledig2017photorealisticsingleimagesuperresolution}, they are prone to hallucinations and training instability~\cite{saad2023surveytrainingchallengesgenerative}. Furthermore, regression and GAN methods are typically deterministic, failing to characterize the posterior distribution required to distinguish genuine anatomy from model artifacts~\cite{hullermeier2021aleatoric}. In this context, Denoising Diffusion Probabilistic Models (DDPMs)~\cite{ho2020denoising} have emerged as a powerful paradigm for medical inverse problems, recover high-frequency details by iteratively refining noise into structured images~\cite{jalal2021robust}. Yet, standard diffusion architectures are designed for image-to-image tasks, making them ill-suited for the high-dimensional, non-local nature of acoustic waveforms.

In this work, we present DiffSOS, a conditional diffusion framework designed for high-fidelity SoS reconstruction. The task is reformulated as a conditional generation process, leveraging a specialized ControlNet architecture~\cite{zhang2023addingControlNet} to extract hierarchical features directly from radiofrequency waveform inputs. To ensure spectral consistency, a hybrid loss function incorporating frequency-domain constraints is employed. Finally, stochastic Denoising Diffusion Implicit Model (DDIM) sampling~\cite{song2020denoising} is utilized to accelerate inference and generate pixel-wise uncertainty maps. In summary, the contributions of this paper are as follows:

\begin{enumerate}
    \item We propose \textbf{DiffSOS}, the first conditional diffusion framework with an acoustic ControlNet to map radiofrequency waveforms directly to SoS maps, bridging the sensor-spatial domain gap and bypassing the FWI.
    \item A spectral consistency loss is introduced that enforces frequency-domain constraints, preserving sharp acoustic boundaries critical for diagnosis.
    \item Near real-time stochastic inference via DDIM sampling achieves superior quality with pixel-wise uncertainty quantification for clinical reliability.
\end{enumerate}
\section{Proposed Method}
\begin{figure}[htbp]
\centering
\includegraphics[width=14cm]{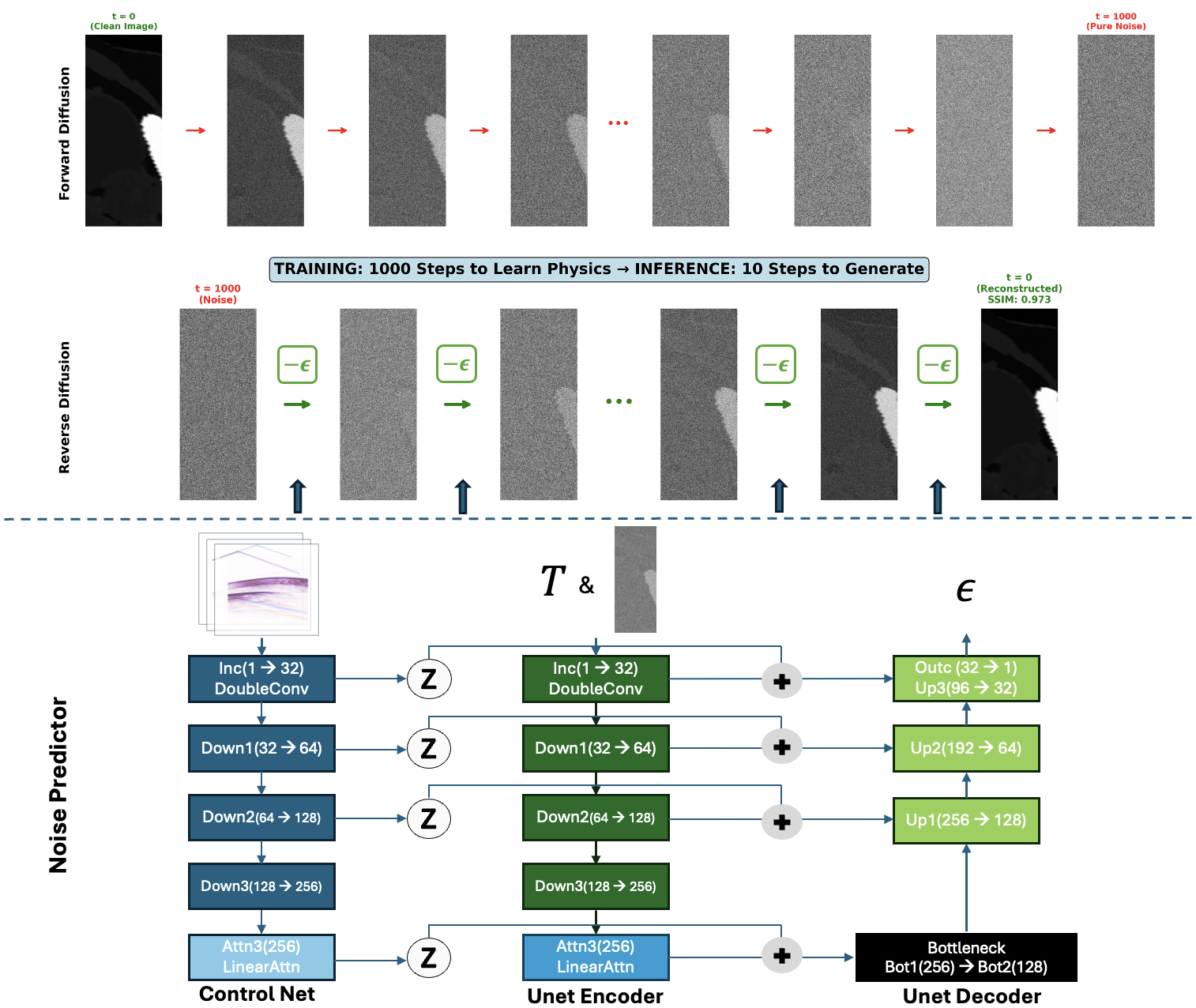}
\caption{Overview of the proposed DiffSOS framework.} \label{method}
\end{figure}

We propose DiffSOS, a conditional diffusion framework shown in Figure \ref{method}, to reconstruct high-fidelity SoS maps $\mathbf{x}_0$ directly from raw acoustic waveforms $\mathbf{y}$. Formulating this as a conditional generative process $p(\mathbf{x}_0 | \mathbf{y})$, we aim to recover complex tissue structures while providing uncertainty estimates.

\subsection{Acoustic ControlNet:} 
The DiffSOS framework is built upon the DDPM backbone~\cite{ho2020denoising}, effectively adapted to maximize information flow from the acoustic to the spatial domain via a specialized ControlNet guidance. The training objective relies on a forward diffusion process that degrades the clean SoS map, $\mathbf{x}_0$, into Gaussian noise, $\mathbf{x}_T$, over $T$ timesteps via a fixed Markov chain:$$q(\mathbf{x}_t | \mathbf{x}_{t-1}) = \mathcal{N}(\mathbf{x}_t; \sqrt{1 - \beta_t}\mathbf{x}_{t-1}, \beta_t \mathbf{I})$$where $\beta_t \in (0, 1)$ represents a pre-defined variance schedule controlling the noise magnitude, and $\mathbf{I}$ denotes the identity matrix. In the reverse generative process, we aim to recover the underlying tissue structure by training a neural network, $f_\theta$, to act explicitly as a noise predictor, approximating the noise component, $\epsilon$, added at each time step based on the conditioning waveform information. To ensure this prediction is strictly grounded in the physical acoustic measurements, a specialized acoustic ControlNet is employed. Instead of simple concatenation, which struggles to bridge the domain gap between 1D sensor data and 2D spatial structures, we use a parallel ControlNet branch that processes the input waveform $\mathbf{y}$. The parallel ControlNet independently processes $\mathbf{y}$ to extract hierarchical features, injecting them into the U-Net encoder via additive coupling. The $i$-th scale feature map $\mathbf{h}_{enc}^{(i)}$ is:
\begin{equation}
    \mathbf{h}_{enc}^{(i)} = \Phi_{enc}^{(i)}(\mathbf{x}_{in}, t) + \mathcal{Z}(\Phi_{ctrl}^{(i)}(\mathbf{y}, t))
\end{equation}
where $\Phi_{enc}$ and $\Phi_{ctrl}$ denote the U-Net and ControlNet neural blocks respectively, and $\mathcal{Z}(\cdot)$ is a zero-initialized $1 \times 1$ convolution. This zero-initialization ensures that the ControlNet does not distort the diffusion priors at the start of training, allowing the model to gradually learn the mapping from acoustic signals to spatial features without instability.

\subsection{Hybrid Multi-objective Loss:}
To ensure structural fidelity and prevent oversmoothing, we design a hybrid objective function:
\begin{equation}
    \mathcal{L}_{total} = \mathcal{L}_{noise} + \lambda_{rec} \mathcal{L}_{rec} + \lambda_{freq} \mathcal{L}_{freq}
\end{equation} 
With he primary noise prediction loss term: $$\mathcal{L}_{noise} = \| \epsilon - f_\theta(\mathbf{x}_{in}, t, \mathbf{y}) \|_2^2$$ It drives the fundamental diffusion mechanism \cite{ho2020denoising}. However, relying solely on noise prediction can lead to hallucinated structures that satisfy the diffusion prior but deviate from the ground truth anatomy. To counteract this, we incorporate a reconstruction consistency loss: $$\mathcal{L}_{rec} = \| \mathbf{x}_0 - \hat{\mathbf{x}}_0 \|_1$$ Here, the estimated clean image $\hat{\mathbf{x}}_0$ is derived analytically using the cumulative product of $(1-\beta_t)$, denoted as $\bar{\alpha}_t$: $\hat{\mathbf{x}}_0 = (\mathbf{x}_t - \sqrt{1 - \bar{\alpha}_t} f_\theta) / \sqrt{\bar{\alpha}_t}$. This term acts as a strong spatial regularizer, enforcing pixel-wise accuracy. Finally, to address the critical issue of spectral bias where models fail to generate high-frequency texture, we introduce the frequency loss: $$\mathcal{L}_{freq} = \| |\mathcal{F}(\epsilon)| - |\mathcal{F}(f_\theta)| \|_1$$ By minimizing the discrepancy in the Fourier amplitude spectra ($\mathcal{F}$) of the predicted noise, the model is forced to explicitly learn the high-frequency components essential for defining sharp tissue boundaries. 

\subsection{Stochastic Inference and Uncertainty:} To facilitate clinical utility, we utilize Denoising Diffusion Implicit Models (DDIM) during inference. This process enables non-Markovian sampling, significantly accelerating the reconstruction without retraining. The update step from $t$ to $t-1$ is defined as:
\begin{equation}
    \mathbf{x}_{t-1} = \sqrt{\bar{\alpha}_{t-1}} \hat{\mathbf{x}}_0 + \sqrt{1 - \bar{\alpha}_{t-1} - \sigma_t^2} \cdot f_\theta(\mathbf{x}_t, t, \mathbf{y}) + \sigma_t \epsilon_t
\end{equation}
where $\hat{\mathbf{x}}_0$ is the predicted clean image and $\epsilon_t \sim \mathcal{N}(0, \mathbf{I})$ is random noise, $\bar{\alpha}_{t-1}$ represents the cumulative noise schedule. To control the stochasticity parameter $\sigma_t = \eta \sqrt{(1-\bar{\alpha}_{t-1})/(1-\bar{\alpha}_t)} \sqrt{1-\bar{\alpha}_t/\bar{\alpha}_{t-1}}$, by setting $\eta > 0$, we retain the probabilistic nature of the model even with fewer steps. This stochasticity enables uncertainty quantification. By running $N$ Monte Carlo inference passes for a single input waveform $\mathbf{y}$, we generate an ensemble of predictions $\{\hat{\mathbf{x}}_0^{(k)}\}_{k=1}^N$. The pixel-wise uncertainty map $\mathbf{U}$ is calculated as the variance of these predictions:
$$
    \mathbf{U}_{i,j} = \text{Var}\left( \{ \hat{\mathbf{x}}_{0, i,j}^{(k)} \}_{k=1}^N \right)
$$.

\section{Experiments and Results}
\subsection{Dataset.} We validate the proposed method on the OpenPros dataset\footnote{\url{https://huggingface.co/datasets/ashynf/OpenPros}}~\cite{wang2025openpros}, a public benchmark for prostate USCT containing anatomically realistic 2D SoS phantoms derived from clinical MRI/CT and ex-vivo ultrasound. Each sample comprises a ground truth SoS map at $400 \times 160$ resolution and corresponding raw acoustic waveforms with dimensions $40 \times 1000 \times 161$ (sources $\times$ time samples $\times$ receivers), generated via Finite-Difference Time-Domain simulation to capture complex wave phenomena. For this study, we utilize a subset of 1,140 paired samples, partitioned with an 8:1:1 ratio for training, validation, and testing.

\subsection{Experimental Setup} Implemented in PyTorch and trained on a single NVIDIA RTX 4090 GPU, DiffSOS requires approximately 22 hours for convergence with 4000 epochs. Batch size is 16 and the Adam optimizer is used with cosine annealing scheduler of $lr_{max} = 8\times10^{-5}$, following a 200-epoch warmup. To address the significant scale disparity observed among loss components, we performed a grid search, finding the best weighting coefficients at $\lambda_{rec} = 0.1$ and $\lambda_{freq} = 0.01$. The diffusion process is configured with $T=1000$ noise steps and a time embedding dimension of 256, while model weights are updated using an exponential moving average with a decay rate of 0.995.

\subsection{Results} To comprehensively assess reconstruction, we employ four quantitative metrics; Multi-Scale Structural Similarity (MS-SSIM) to evaluate structural integrity across scales; Peak Signal-to-Noise Ratio (PSNR) for pixel-level fidelity; Mean Absolute Error (MAE) to quantify the physical accuracy of velocity values; and Pratt's Figure of Merit (FOM)~\cite{pratt2001digital} to assess edge preservation and the sharpness of acoustic boundaries. The optimal model checkpoint is selected based on the highest MS-SSIM achieved on the validation set, and all final performance metrics are reported on the held-out test set, utilizing $T=10$ inference steps via DDIM sampling.

\subsubsection{Model Comparison} We benchmark DiffSOS, with approximately 8.7M trainable parameters, against InversionNet~\cite{Wu_2020} and VelocityGAN~\cite{zhang2019datadrivenseismicwaveforminversion}, which utilize 16.2M and 17.3M trainable parameters, respectively. To isolate the benefits of the diffusion process, we evaluate a custom conditional GAN (cGAN)~\cite{mirza2014conditionalgenerativeadversarialnets,pix2pix2017} baseline of 11.2M trainable parameters that utilizes our proposed acoustic ControlNet architecture and loss functions within an adversarial framework. 
\begin{figure}
\centering
\includegraphics[width=15cm]{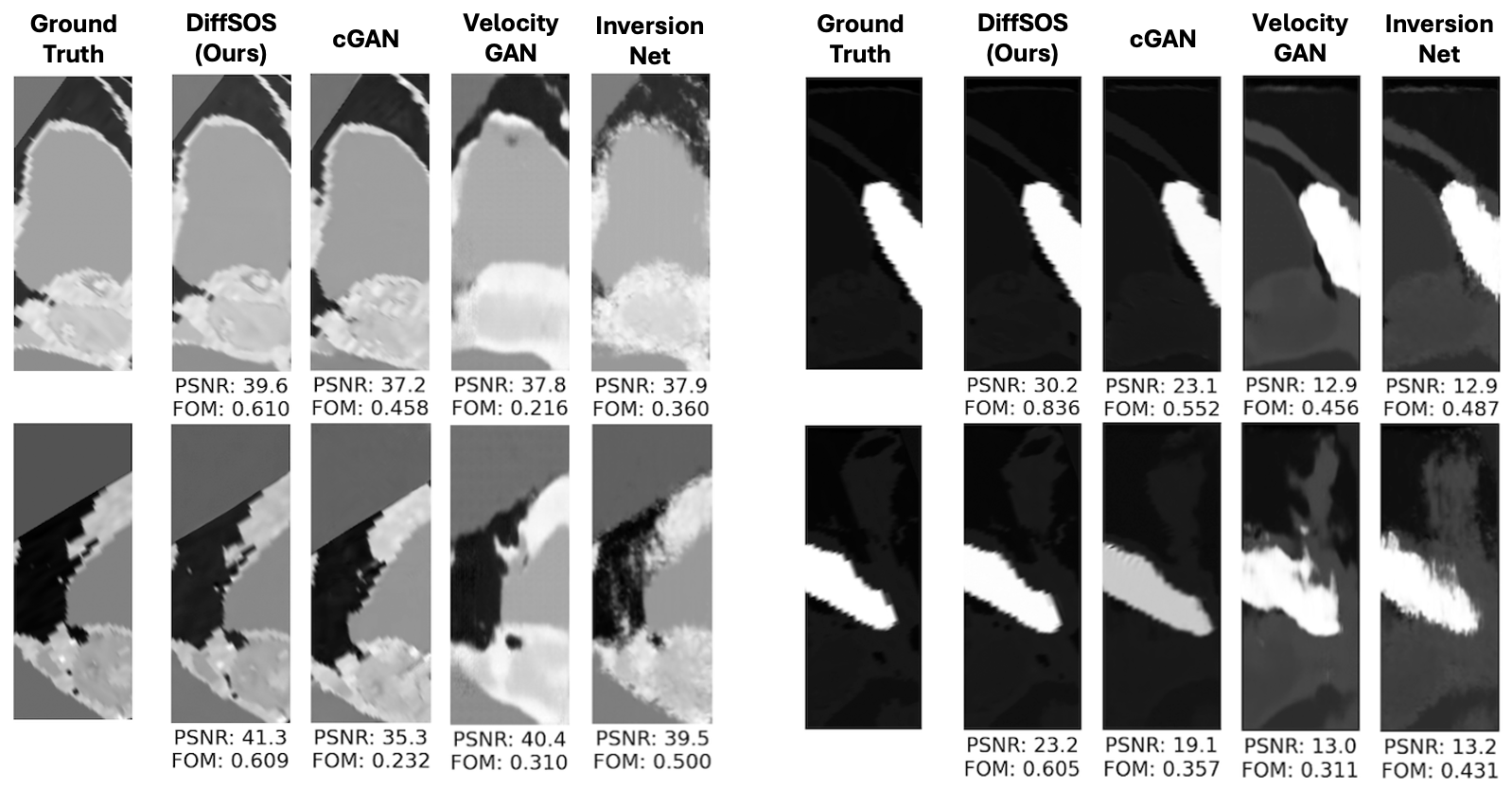}
\caption{Qualitative comparison with baseline methods.} \label{fig:modelComp}
\end{figure}

All models are trained and evaluated on identical data splits. Qualitative results are presented in Figure \ref{fig:modelComp}. Deterministic baselines such as InversionNet and VelocityGAN exhibit characteristic oversmoothing and fail to resolve high-frequency anatomical boundaries due to regression to the mean. While the cGAN baseline recovers high-frequency textures, it struggles with structural accuracy and often hallucinates details absent in the ground truth. In contrast, DiffSOS effectively prevents oversmoothing while maintaining structural fidelity, reconstructing fine-grained tissue heterogeneity with high precision. Quantitative results are shown in Table \ref{tab:comparison}, where our method consistently outperforms baselines across all metrics.

\begin{table}[h!]
\caption{Quantitative comparison of reconstruction methods. Best results in \textbf{bold}. \textbf{$\pm$} denotes standard deviation.}\label{tab:comparison}
\centering
\begin{tabular}{|l|c|c|c|c|}
\hline
Method & MS-SSIM $\uparrow$ & PSNR (dB) $\uparrow$ & MAE $\downarrow$ & FOM $\uparrow$ \\
\hline
InversionNet & 0.844 ± 0.091 & 21.70 ± 10.90 & 0.233 ± 0.180 & 0.412 ± 0.081 \\
VelocityGAN & 0.849 ± 0.085 & 21.86 ± 10.91 & 0.229 ± 0.177 & 0.336 ± 0.080 \\
cGAN & 0.919 ± 0.031 & 27.37 ± 6.26 & 0.069 ± 0.034 & 0.436 ± 0.110 \\
\textbf{DiffSOS(Ours)} & \textbf{0.957 ± 0.031} & \textbf{30.17 ± 8.09} & \textbf{0.048 ± 0.035} &  \textbf{0.657 ± 0.127} \\
\hline
\end{tabular}
\end{table}

\subsubsection{Ablation Studies}
 We validate the effectiveness of our architectural design and loss components in Table~\ref{tab:ablation}. First, we compare against a Concatenation-Only variant, in which the radiofrequency waveform is concatenated with the 2D noisy latent $\mathbf{x}_t$ at the U-Net input. This approach yields poor performance of MS-SSIM 0.718 due to severe conditioning failure; unable to bridge the sensor-spatial domain gap, the model generates random structures unrelated to the input waveform physics. We also compared against a Cross-Attention variant, and the waveform is injected via Cross-Attention layers at each decoder block. This approach achieves MS-SSIM 0.711, as the global attention operation struggles to preserve fine-grained spatial correspondence. In contrast, our acoustic ControlNet successfully enforces physical constraints, boosting MS-SSIM to 0.957.
 
\begin{table}[h!]
\caption{Ablation study: Quantitative comparison of different model variants. Best results are shown in \textbf{bold}. \textbf{$\pm$} denotes standard deviation.}\label{tab:ablation}
\centering
\begin{tabular}{|l|c|c|c|c|}
\hline
Model Variant & MS-SSIM $\uparrow$ & PSNR (dB) $\uparrow$ & MAE $\downarrow$ & FOM $\uparrow$ \\
\hline
Concatenation Only & 0.718 ± 0.054 & 12.37 ± 2.01 & 0.375 ± 0.105 & 0.133 ± 0.025\\
Cross-Attention & 0.711 ± 0.011 & 14.98 ± 1.68 & 0.378 ± 0.023 & 0.027 ± 0.017 \\ 
Baseline  & 0.944 ± 0.046 & 29.11 ± 8.57 & 0.059 ± 0.045 & 0.621 ± 0.118 \\
Baseline+Freq   & 0.927 ± 0.043 & 26.87 ± 6.88 & 0.119 ± 0.071 & 0.601 ± 0.116 \\
Baseline+L1  & 0.951 ± 0.040 & 29.80 ± 8.82 & 0.054 ± 0.042 & 0.620 ± 0.127\\

\textbf{Full (Proposed)} & \textbf{0.957 ± 0.031} & \textbf{30.17 ± 8.09} & \textbf{0.048 ± 0.035} &  \textbf{0.657 ± 0.127}\\
\hline
\end{tabular}
\end{table}

\begin{figure}[h!]
\centering
\includegraphics[width=14cm]{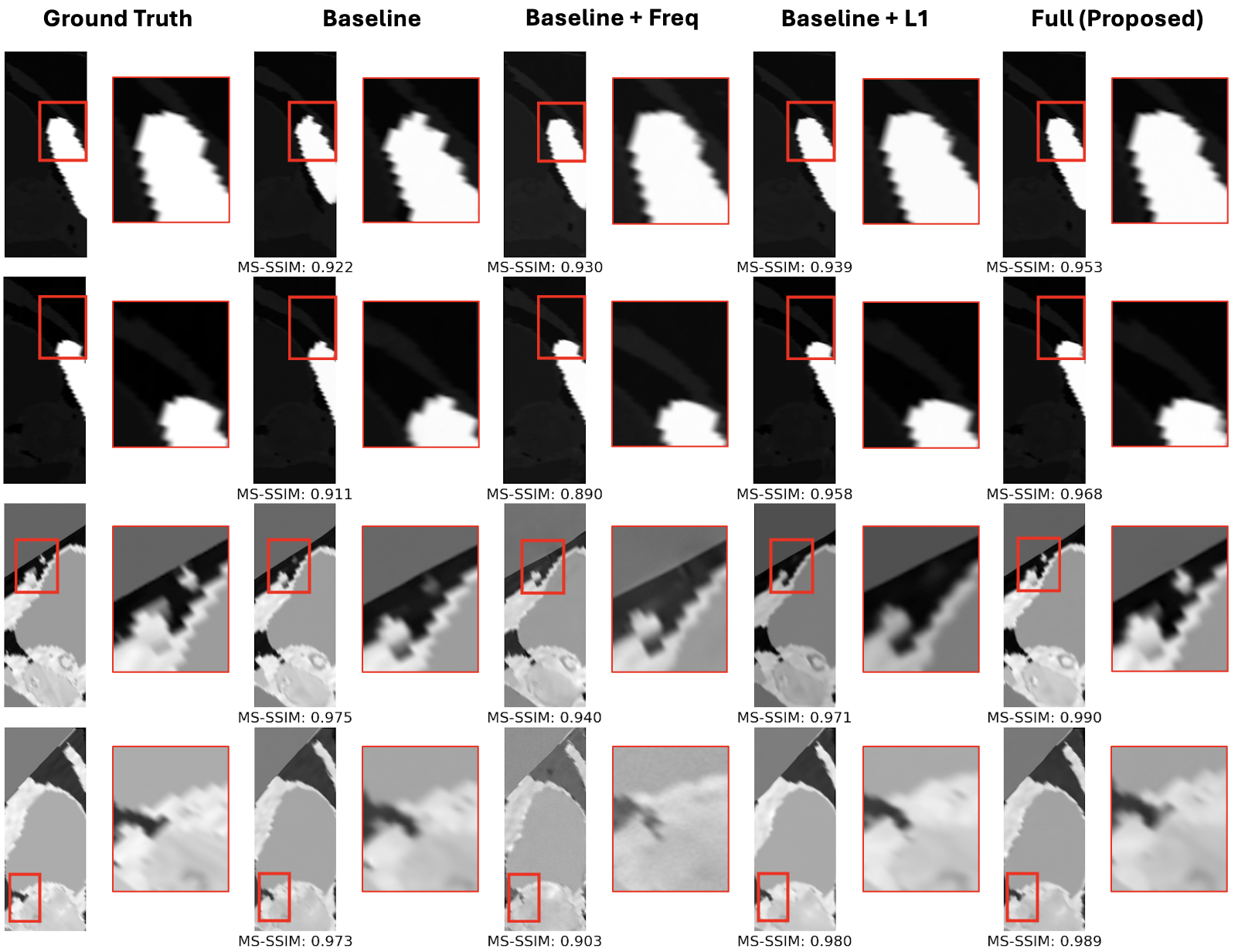}
\caption{Visual ablation study of loss components.} \label{fig:ablation}
\end{figure}

We further analyze loss contributions relative to the DDPM baseline. Incorporating $\mathcal{L}_{rec}$ provides a direct physical constraint on velocity values, yielding consistent improvements across MS-SSIM, PSNR, and MAE. Notably, the FOM does not increase until we include the frequency loss. However, adding $\mathcal{L}_{freq}$ in isolation degrades performance, causing a sharp increase in MAE, as the model prioritizes spectral matching over spatial coherence, introducing pixel-level distortions to satisfy frequency constraints. The full model achieves optimal performance by combining both terms. $\mathcal{L}_{rec}$ acts as a spatial anchor, enabling $\mathcal{L}_{freq}$ to sharpen high-frequency boundaries without distortion. As shown in Figure \ref{fig:ablation}, this synergy yields physically plausible reconstructions characterized by precise edge definition and accurate structural details.

\subsubsection{Uncertainty}
\begin{figure}[h!]
\centering
\includegraphics[width=16.5cm]{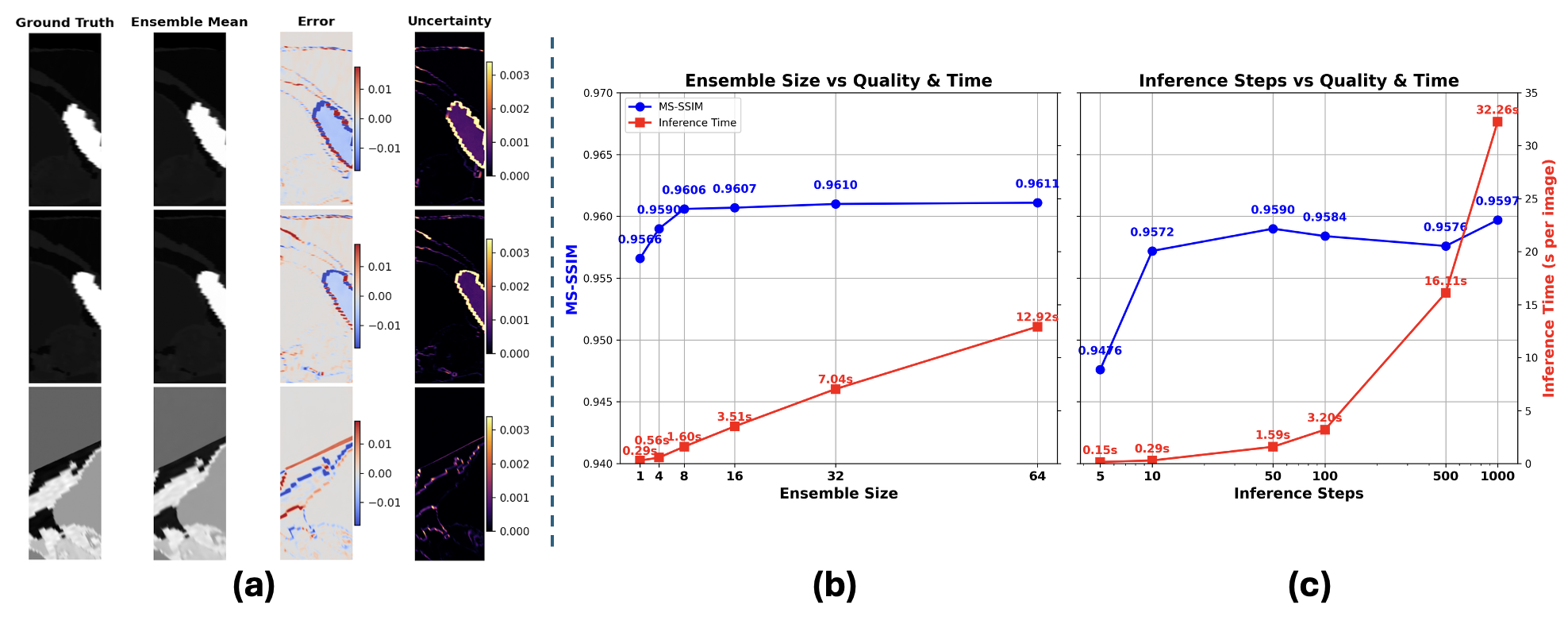}
\caption{\textbf{Uncertainty and Efficiency Analysis.} \textbf{(a)} Uncertainty maps highly correlate with reconstruction errors. Quantitative trade-offs for ensemble size \textbf{(b)} and sampling steps \textbf{(c)} confirm that high-fidelity reconstruction is achievable within clinically viable timeframes.} \label{fig:uq_inference}
\end{figure}

Beyond reconstruction accuracy, assessing model confidence is essential for clinical decision-making. We leverage the stochastic nature of our DDIM sampler to quantify aleatoric uncertainty, the inherent variability arising from the generative sampling process. We execute $N=10$ Monte Carlo inference passes using fixed model weights but different input random noise for the reverse diffusion process, computing a pixel-wise variance map across these predictions. This metric captures the reconstruction's sensitivity to sampling noise. As visualized in Figure \ref{fig:uq_inference}(a), the generated uncertainty maps act as a reliable proxy for failure, with high-variance regions strongly correlating with large reconstruction errors, defined here as the difference between ground truth and ensemble mean. Moreover, as demonstrated in Figure \ref{fig:uq_inference}(b), this ensemble strategy remains computationally efficient, enabling rapid confidence assessment in clinical workflows, allowing clinicians to distinguish high-confidence anatomical structures from potential artifacts.

\subsubsection{Efficiency\label{uq_and_time}}
We analyze the computational efficiency of DiffSOS to validate its clinical feasibility. Standard diffusion models typically rely on a long Markov chain to generate high-quality outputs, which is computationally prohibitive for real-time applications. As shown in Figure \ref{fig:uq_inference}(c), utilizing stochastic DDIM sampling with a reduced trajectory of $T=10$ steps slashes inference time from 32.26s ($T=1000$) to 0.29s per image, achieving a speedup of over two orders of magnitude with negligible quality loss. The supplementary video further demonstrates this consistency, showing that global structures are resolved even at $T=10$. This confirms that DiffSOS enables near real-time reconstruction without compromising structural fidelity. As a secondary observation, Figure \ref{fig:uq_inference}(b) demonstrates that the ensemble strategy further stabilizes performance, marginally increasing MS-SSIM to 0.9611 for high-precision scenarios.

\section{Discussion and Conclusion}
We introduce \textbf{DiffSOS}, a conditional diffusion framework that reconstructs high-fidelity Speed-of-Sound maps directly from radiofrequency acoustic waveforms. By integrating the acoustic ControlNet with a spectral consistency loss, our method overcomes the sensor-spatial domain gap, avoiding the oversmoothing typical of deterministic baselines while eliminating the computational burden of iterative FWI. Furthermore, accelerated DDIM sampling facilitates near real-time inference and provides pixel-wise uncertainty maps for reliable clinical decision-making. Future efforts will explore diverse acquisition geometries, such as sparse waveform configurations, and adapt the framework to new clinical domains like breast USCT, while extending the model to jointly reconstruct acoustic attenuation alongside Speed-of-Sound maps for comprehensive tissue characterization.

\newpage
\bibliographystyle{unsrt}  
\bibliography{reference}  

@article{Wu_2020,
   title={InversionNet: An Efficient and Accurate Data-Driven Full Waveform Inversion},
   volume={6},
   ISSN={2573-0436},
   url={http://dx.doi.org/10.1109/TCI.2019.2956866},
   DOI={10.1109/tci.2019.2956866},
   journal={IEEE Transactions on Computational Imaging},
   publisher={Institute of Electrical and Electronics Engineers (IEEE)},
   author={Wu, Yue and Lin, Youzuo},
   year={2020},
   pages={419–433} }

@INPROCEEDINGS{10793606,
  author={Yan, Yan and Bustamante, David and Jin, Gaofei and Spatarelu, Catalina and Luke, Geoffrey and Mehrmohammadi, Mohammad},
  booktitle={2024 IEEE Ultrasonics, Ferroelectrics, and Frequency Control Joint Symposium (UFFC-JS)}, 
  title={Ultrasound Tomography with a Ring array Ultrasound Transducer for Image-Guided Activation of Phase-Change Nanodroplets}, 
  year={2024},
  volume={},
  number={},
  pages={1-4},
  doi={10.1109/UFFC-JS60046.2024.10793606}}

@article{wang2025openpros,
  title={OpenPros: A Large-Scale Dataset for Limited View Prostate Ultrasound Computed Tomography},
  author={Wang, Hanyu and Wu, Yifan and Feng, Yuxuan and Jin, Peng and Feng, Shaoyuan and Mao, Yiming and Wiskin, James and Turkbey, Baris and Pinto, Peter A and Wood, Bradford J and Luo, Sidong},
  journal={arXiv preprint arXiv:2505.12261},
  year={2025}
}

@article{rau2021speed,
  title={Speed-of-sound imaging using diverging waves},
  author={Rau, Richard and Schweizer, Dieter and Vishnevskiy, Valery and Goksel, Orcun},
  journal={International Journal of Computer Assisted Radiology and Surgery},
  volume={16},
  number={7},
  pages={1201--1211},
  year={2021},
  month={jul},
  publisher={Springer},
  doi={10.1007/s11548-021-02426-w},
  pmid={34160749}
}

@misc{zhang2019datadrivenseismicwaveforminversion,
      title={Data-driven Seismic Waveform Inversion: A Study on the Robustness and Generalization}, 
      author={Zhongping Zhang and Youzuo Lin},
      year={2019},
      eprint={1809.10262},
      archivePrefix={arXiv},
      primaryClass={eess.SP},
      url={https://arxiv.org/abs/1809.10262}, 
}

@article{tarantola1984inversion,
  title={Inversion of seismic reflection data in the acoustic approximation},
  author={Tarantola, Albert},
  journal={Geophysics},
  volume={49},
  number={8},
  pages={1259--1266},
  year={1984},
  publisher={Society of Exploration Geophysicists},
  doi={10.1190/1.1441754},
  url={https://doi.org/10.1190/1.1441754}
}

@article{virieux2009overview,
  title={An overview of full-waveform inversion in exploration geophysics},
  author={Virieux, Jean and Operto, St{\'e}phane},
  journal={Geophysics},
  volume={74},
  number={6},
  pages={WCC1--WCC26},
  year={2009},
  publisher={Society of Exploration Geophysicists},
  doi={10.1190/1.3238367},
  url={https://doi.org/10.1190/1.3238367}
}

@article{byrd1995limited,
  title={A limited memory algorithm for bound constrained optimization},
  author={Byrd, Richard H and Lu, Peihuang and Nocedal, Jorge and Zhu, Ciyou},
  journal={SIAM Journal on Scientific Computing},
  volume={16},
  number={5},
  pages={1190--1208},
  year={1995},
  publisher={SIAM},
  doi={10.1137/0916069},
  url={https://doi.org/10.1137/0916069}
}

@misc{lee2024systematicbiasmachinelearning,
      title={A Systematic Bias of Machine Learning Regression Models and Its Correction: an Application to Imaging-based Brain Age Prediction}, 
      author={Hwiyoung Lee and Shuo Chen},
      year={2024},
      eprint={2405.15950},
      archivePrefix={arXiv},
      primaryClass={stat.ML},
      url={https://arxiv.org/abs/2405.15950}, 
}

@misc{ledig2017photorealisticsingleimagesuperresolution,
      title={Photo-Realistic Single Image Super-Resolution Using a Generative Adversarial Network}, 
      author={Christian Ledig and Lucas Theis and Ferenc Huszar and Jose Caballero and Andrew Cunningham and Alejandro Acosta and Andrew Aitken and Alykhan Tejani and Johannes Totz and Zehan Wang and Wenzhe Shi},
      year={2017},
      eprint={1609.04802},
      archivePrefix={arXiv},
      primaryClass={cs.CV},
      url={https://arxiv.org/abs/1609.04802}, 
}

@ARTICLE{7797130,
  author={Zhao, Hang and Gallo, Orazio and Frosio, Iuri and Kautz, Jan},
  journal={IEEE Transactions on Computational Imaging}, 
  title={Loss Functions for Image Restoration With Neural Networks}, 
  year={2017},
  volume={3},
  number={1},
  pages={47-57},
  doi={10.1109/TCI.2016.2644865}}

@misc{saad2023surveytrainingchallengesgenerative,
      title={A Survey on Training Challenges in Generative Adversarial Networks for Biomedical Image Analysis}, 
      author={Muhammad Muneeb Saad and Ruairi O'Reilly and Mubashir Husain Rehmani},
      year={2023},
      eprint={2201.07646},
      archivePrefix={arXiv},
      primaryClass={cs.LG},
      url={https://arxiv.org/abs/2201.07646}, 
}

@INPROCEEDINGS{tof,
  author={Liu, Zhaohui and Wang, Jiameng and Ding, Mingyue and Yuchi, Ming},
  booktitle={2021 IEEE International Ultrasonics Symposium (IUS)}, 
  title={Deep Learning Ultrasound Computed Tomography with Sparse Transmissions}, 
  year={2021},
  volume={},
  number={},
  pages={1-4},
  doi={10.1109/IUS52206.2021.9593459}}

@article{liu2025reconstruction,
  title={Reconstruction of reflection ultrasound computed tomography with sparse transmissions using conditional generative adversarial network},
  author={Liu, Zhaohui and Zhou, Xiang and Yang, Hantao and Zhang, Qiude and Zhou, Liang and Wu, Yun and Liu, Quanquan and Yan, Weicheng and Song, Junjie and Ding, Mingyue and Yuchi, Ming and Qiu, Wu},
  journal={Ultrasonics},
  volume={145},
  pages={107486},
  year={2025},
  publisher={Elsevier},
  doi={10.1016/j.ultras.2024.107486},
  url={https://doi.org/10.1016/j.ultras.2024.107486}
}

@article{ho2020denoising,
    title={Denoising Diffusion Probabilistic Models},
    author={Jonathan Ho and Ajay Jain and Pieter Abbeel},
    year={2020},
    journal={arXiv preprint arxiv:2006.11239}
}

@article{song2020denoising,
  title={Denoising Diffusion Implicit Models},
  author={Song, Jiaming and Meng, Chenlin and Ermon, Stefano},
  journal={arXiv:2010.02502},
  year={2020},
  month={October},
  abbr={Preprint},
  url={https://arxiv.org/abs/2010.02502}
}

@book{pratt2001digital,
  title={Digital image processing: PIKS inside},
  author={Pratt, William K},
  year={2001},
  publisher={John Wiley \& Sons}
}

@misc{zhang2023addingControlNet,
  title={Adding Conditional Control to Text-to-Image Diffusion Models}, 
  author={Lvmin Zhang and Anyi Rao and Maneesh Agrawala},
  booktitle={IEEE International Conference on Computer Vision (ICCV)},
  year={2023},
}

@misc{mirza2014conditionalgenerativeadversarialnets,
      title={Conditional Generative Adversarial Nets}, 
      author={Mehdi Mirza and Simon Osindero},
      year={2014},
      eprint={1411.1784},
      archivePrefix={arXiv},
      primaryClass={cs.LG},
      url={https://arxiv.org/abs/1411.1784}, 
}

@article{pix2pix2017,
  title={Image-to-Image Translation with Conditional Adversarial Networks},
  author={Isola, Phillip and Zhu, Jun-Yan and Zhou, Tinghui and Efros, Alexei A},
  journal={CVPR},
  year={2017}
}

@article{hullermeier2021aleatoric,
  title={Aleatoric and epistemic uncertainty in machine learning: {An} introduction to concepts and methods},
  author={H{\"u}llermeier, Eyke and Waegeman, Willem},
  journal={Machine Learning},
  volume={110},
  number={3},
  pages={457--506},
  year={2021},
  publisher={Springer},
  doi={10.1007/s10994-021-05946-3}
}

@article{jalal2021robust,
  title={Robust Compressed Sensing MRI with Deep Generative Priors},
  author={Jalal, Ajil and Arvinte, Marius and Daras, Giannis and Price, Eric and Dimakis, Alexandros G and Tamir, Jonathan I},
  journal={Advances in Neural Information Processing Systems},
  year={2021}
}






\end{document}